# A Novel Approach in detecting pose orientation of a 3D face required for face registration

Parama Bagchi[#1], Debotosh Bhattacharjee[*2], Mita Nasipuri[*3], Dipak Kumar Basu[*4]

[#1]Dept. of CSE, MCKV Institute of Engineering, Kolkata 711204, India.
[1]paramabagchi@gmail.com
[*2] Dept. of CSE, Jadavpur University, Kolkata 700032, India
[2] debotosh@ieee.org
[*3]Dept. of CSE, Jadavpur University, Kolkata 700032, India
[3] mitanasipuri@gmail.com
[*4]Former Professor, Dept. of CSE, Jadavpur University, Kolkata 700032, India
[4] dipakkbasu@gmail.com

*Abstract*— **In this paper we present a novel approach that takes as input a 3D image and gives as output its pose i.e. it tells whether the face is oriented with respect the X, Y or Z axes with angles of rotation up to 40º. All the experiments have been performed on the FRAV3D Database. After applying the proposed algorithm to the 3D facial surface we have obtained i.e. on 848 3D face images our method detected the pose correctly for 566 face images ,thus giving an approximately 67 % of correct pose detection.**

## I. INTRODUCTION

3D face registration has been a crucial and an important field of study in 3D face recognition. In 3D face recognition, registration is a key pre-processing step. Registration is actually the step in 3D modelling to find correspondences between two 3D views. The main purpose is to derive a rigid transformation (namely the rotation matrix and the translation vector in the case of rigid objects) that will align the views in a common coordinate system. Registration approaches use the distance between the aligned facial surfaces as a measure of how well faces match. To align the 3D facial shapes, nearly all state-of-the-art 3D face recognition methods minimize the distance between two face shapes or between a face shape and an average face model. Normally, registration is done on the basis of some landmarks or by aligning to some intrinsic coordinate system defined on the basis of those landmarks. Much of the 3D face work presented in the literature uses low noise 3D data in a frontal pose and normalization techniques that do not have good performances in case of different pose variations. In contrast, our method requires us to be able to identify and analyze the pose orientation of any 3D face image across all varied poses thus enabling an effective way of face registration. In Section II, a description of our method for pose analysis has been described. In Section III, a comparative analysis of the proposed method over other pre-existing methods of registration as well as the complexity analysis of the present method has been discussed. Experimental results are discussed in Section IV and finally in the Section V, the conclusions are enlisted. In this section, we are going to discuss some related works done in the field of 3D registration more specifically in the field of pose analysis and thus compare how our method of pose orientation outperforms the already prevalent methods of pose orientation. In [1], an analysis of pose variation has been presented but it has not been mentioned about how the estimation of pose orientations with respect to X, Y and Z axes has been estimated. In [2], the facial symmetry is dealt with in determining the pose of 3D faces. In the present work, the authors have proposed an effective method to estimate the symmetry plane of facial surface. But no discussions on how to find the various pose orientations have been highlighted. In [3] also, a robust approach on 3D registration has been done but no specific discussions about the orientation of the 3D face with respect to the X,Y and Z axes has been highlighted. In [4], the authors have used profile based symmetry assuming an image to be already aligned with respect to X, Y and Z axes but no discussions of how to retrieve the pose alignment with respect to the various axes have been highlighted. In [5], a discussion of pose has been given with respect to X,Y and Z axes but more specifically the Z corrected poses have been specifically discussed with no emphasis on how to detect orientations across X and Y automatically. Some other pieces of work include the ones done by Maurer [6] on 3D recognition by extracting 3D texture. Another work was done by Mian[7] on 3D spherical face representation (SFR) used in conjunction with the scale-invariant feature transform (SIFT) descriptor to form a rejection classifier, which quickly eliminates a large number of candidate faces at an early stage for recognition in case of large galleries. This approach automatically segments the eyes, forehead, and the nose regions, which are relatively less sensitive to expressions and matches them separately using a modified iterative closest point (ICP) algorithm.

Another work was done by Queirolo[8] which presented a novel automatic framework to perform 3D face recognition. The proposed method uses a Simulated Annealing-based approach (SA) for range image registration with the Surface Interpenetration Measure (SIM), as similarity measure, in order to match two face images. Another work done by Faltemier [9] is based on 3D face recognition with no discussion on pose variances. In all these works, no emphasis about how to find out the pose alignment of any 3D face being input to the system is discussed. In contrast to all the above methods, our method of pose orientation uses some feature points which could tell exactly in what orientation a 3D face is rotated , thereby increasing the efficiency of registering the input 3D unregistered 3D face model. There can be one-to all registration, registration to a face model, or registration to an intrinsic coordinate system. Our method of registration technique is a one to all registration technique. In this paper we have used a geometrical model in determining the pose orientation of a 3D face which would be described in Section II. All the experiments have been done on the entire FRAV3D database [10]. Fig-1 shows a brief description of our registration system.

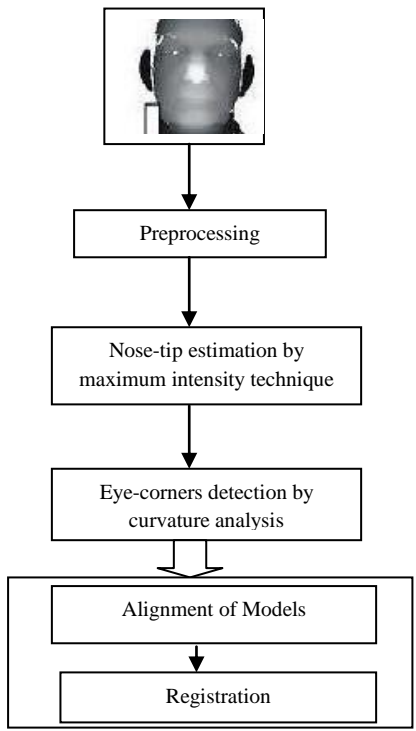

Fig 1. An overview of our proposed system

## II. OVERVIEW OF OUR PROPOSED SYSTEM

The present system for localization of nose-tip and eye-corners consists of the following four steps:-

- Pre-processing
- Nose-tip estimation by maximum intensity technique
- Eye-corners detection by curvature analysis
- Alignment of Models
- Registration

### A. Preprocessing

First of all the 3D face of size [100 by 100] is cropped to dimensions of size [15 70]. Next, the image is thresholded by Otsu's thresholding [11] algorithm. After thresholding the 3D images obtained are shown as in Fig-2.

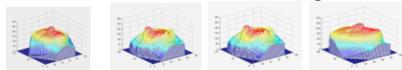

(a) Frontal pose (b) Rotated about y-axis (c) Rotated about x-axis (d) Rotated about z axis
Fig.2. Mesh-grids after thresholding

In the next step the range image is smoothed by Gaussian filter. After smoothing the 3D images obtained are as shown in Fig-3.

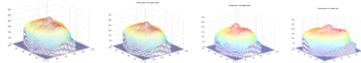

Fig 3. Mesh grids after smoothing

### B. Nose-tip estimation using maximum-intensity Technique:-

For the nose tip localization we have used the maximum intensity concept as the tool for the selection process. Each of the faces (including rotation in any direction in 3D space namely about x-axis, y-axis and z-axis) from the FRAV3D database has been used for localization of nose-tip. A set of points are extracted from both frontal and various poses of face images using a maximum intensity algorithm. The region of interest i.e. the nose tip has been marked on the surface so generated. The Fig 4 shows the 3D faces with nose tip localized. The max-intensity [16] is given below:-

Function Find_Maximum_Intensity (Image)
   Step1:- Set max to 0
   Step 2:- Run loop for I from 1 to width (Image)
   Step 3:- Run loop for J from 1 to height (Image)
   Step 4:- Set val to sum (image (I-1: I+1, J-1: J+1))
   Step 5:- Check if val is greater than max.
   Step 6:- Set val to val2
   Step 7: - End if
   Step 8:- End loop for
   Step 9:- End loop for J
End Function

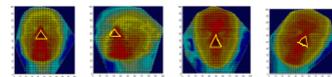

Fig 4. Surface Generation with nose localized

### C. Eye-corners detection by Curvature Analysis:

In this section, we have generated the eye-corners using curvature analysis [13, 14]. In Fig 5 we show an output of our method for detection of eye-corners using curvature analyses in frontal pose.

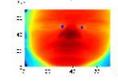

Fig 5. A sample output in frontal pose for detecting eye-corners

After running the algorithm the output obtained is as shown:-

   Row  Col  Curvature
   51   29   0.000410
   50   49   0.000225

The points of highest curvature values are the inner corners of the eye-region. Similarly, we show a sample output in Fig 6 for a 3D image rotated about y-axis and it's corresponding output.

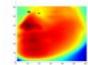

Fig 6. A sample output for an image in rotated pose rotated about y axis for detecting eye–corners

   Row  Col  Curvature
   53   20   0.000998
   53   8    0.000336

Similarly, we show a sample output in Fig 7 for a 3D image rotated about X-axis and it's corresponding output.

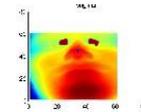

Fig 7. A sample output for an image in rotated pose rotated about x axis for detecting eye–corners

   Row  Col  Curvature
   50   29   0.092934
   51   51   0.0011

Similarly, we show a sample output in Fig 8 for a 3D image rotated about Z-axis and it's corresponding output.

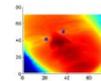

Fig 8. A sample output for an image in rotated pose rotated about z axis for detecting eye–corners

   Row  Col  Curvature
   51   37   0.000357
   43   18   0.000184

### D. Alignment of Models:-

In this section[17], we shall discuss about the alignment and pose estimation of our 3D face images. Fig 9 shows a triangulated face mesh.

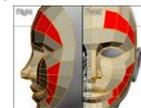

Fig 9. A triangulated face image [12]

From the Figs 5,6,7,8 and 9 we can infer that:-

- If a 3D face is rotated with respect to x-axis, both the eyes must lie on the same horizontal line. Deviation errors are possible but definitely within a minimum threshold.

- If a 3D face is rotated with respect to z-axis, both the eyes must not lie on the same horizontal line.

- If a 3D face is rotated with respect to y-axis, both the eyes must lie on the same horizontal line. Deviation errors are possible but definitely within a minimum threshold. So, as stated above, we have tried to segregate the x-axis and y-axis from z-axis.

Now, we propose a method to segregate the x and y axes:-
- If a 3D face is rotated with respect to x-axis, deviation of nose-tips with respect to y-axes will be more than x axes.
- If a 3D face is rotated with respect to y-axis, then deviation of nose-tips with respect to x-axes will be more than y axes.

Now as stated above, we have tried to segregate the x-axis and y-axis. Now our discussions above are written in the form of an algorithm.

**Algorithm 1:-**
**Input:- 3D face Image**
**Output: - Pose Alignment returned with respect to X, Y and Z axes**
Step 1:- From the FRAV3D face gallery of a person, choose any frontal image of any person.
Step 2:- Find out the nose-tip of the person with a frontal pose by maximum intensity technique.
Step 3:- Find out the eye-corners of the person with a frontal pose by curvature analysis
Step 4:- Find out the nose-tip of the person with the rotated pose by maximum intensity technique.
Step 5:- Find out the eye-corners of the person with the rotated pose by curvature analysis [13, 14].
Step 6:- Assign the x coordinates of the eye-corners of the rotated pose to x1coordf and x2coodrf.
Step 7:- Also assign the xcoordinates of the nose-tips of frontal and rotated pose to variables x1 and x2.
Step 8:- Also assign the ycoordinates of the nose-tips of frontal and rotated pose to variables y1 and y2.
Step 9:- Assign diff = x1coordf - x2coordf.
Step 10:- Check if diff is greater than ℮ then
Step 11:- Print 3D face is rotated with respect to Z axis
Step 12:- Else
Step 13:- Check if x1>x2 or x1<x2
Step 14:- Check if abs(x1 - x2) >= abs (y1 - y2)
Step 15:- Print 3D face rotated with respect to Y axis
Step 16:- End if
Step 17:- End if
Step 18:- Check if y1>y2 or y1<y2
Step 19:- Check if abs (y1-y2)>=abs(x1-x2)
Step 20:- Print 3D face rotated with respect to X axis
Step 21:- End if
Step 22:- End if
Step 23:- End if
Step 24:- End of Algorithm

Here we note that, the value of the parameter ℮ in our algorithm is set to 2, because if we notice the Figs 5, 6, 7, 8 and 9 the eye-corners in case of X and Y axes vary at the most by a degree of 2. Hence keeping that in mind we have set the value of ℮ to 2 because eyes are normally at the same level. We haven't taken the depth into consideration because depth may vary according to pose changes so that parameter cannot be taken into consideration.

### III. COMPARATIVE ANALYSIS AND ALGORITHMIC COMPLEXITY OF THE PROPOSED TECHNIQUE

In this section, we are going to first describe the performance analysis of the max intensity technique[15]. This technique can only detect the nose-tip which is not sufficient to calculate the pose-orientation of any 3D rotated face across X,Y and Z axes. In [14], an analysis of curvature of surfaces has been discussed and their consequent uses in 3D face registration and recognition. But, no discussions regarding pose analysis has been done. Also the alignment of models done in [17], has only discussed the means to calculate the translational parameters and rotational parameters. No discussion on how to calculate the pose-orientations has been dealt with. In the present proposed technique, we have, in details discussed the pose-analysis that is required for registration of any 3D face. The reason being that, we have to calculate the translational and rotational parameters on the basis of this pose orientation. The complexity analysis of the 3 methods and our proposed method has been listed below:-
1. Maximum-intensity technique [15]:- $O(n^2)$
2. Curvature-Analysis [14]:- $O(n^2)$
3. Alignment of Models [17]:- $O(n)$ if only translational and scaling parameters are concerned.
4. Alignment of models by ICP algorithm[18]:- $O(n^2)$
5. The present proposed method:- $O(n^2)+ O(n^2)+O(n) = O(n^2)$.

Henceforth, the proposed method depicts that, taking a cumulative analysis of all the methods taken together the complexity of the present methods becomes $O(n^2)$ which is indeed a long processing time but the advantage of the present method over other methods is that, to calculate the correct translational and scaling parameters, it is indeed important to know the pose-orientation of the 3D face. Also for complex rotations, i.e. across XY, YZ axes two or more rotational parameters might be involved. In those cases, calculation of pose-orientation is very much needed.

### IV. EXPERIMENTAL RESULTS

In this section, we present an analysis of how our proposed method for alignment analysis works on the FRAV3D database. The Fig 10 shows some samples taken from the FRAV3D database for images rotated about X-axis.

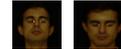

Fig 10. Some samples rotated about x axis from FRAV3D Database

The results of pose –alignment across X axis are shown in Table I.

TABLE I.

| Table 1 | Detection of Pose Alignment across X axes | | |
|---|---|---|---|
| | A | B | C |
| 1 | +5 | 70 | 48 |
| 2 | -5 | 70 | 50 |
| 3 | +18 | 70 | 50 |
| 4 | -18 | 70 | 48 |
| 5 | +40 | 72 | 47 |
| 6 | -40 | 72 | 46 |

Here the notations A, B, C stands for:-
A: - Angles of Alignment with respect to X axis from FRAV3D database
B: - No of 3D images aligned against X-axes
C: - No of alignments detected correctly by our algorithm
The fig 11 shows some samples taken from the FRAV3D database for images rotated about Y-axis.

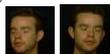

Fig 11. Some samples rotated about y axis from FRAV3D Database

The results of pose –alignment across Y axis are shown in Table II.

TABLE II.

| Table 2 | Detection of Pose Alignment across Y axes | | |
|---|---|---|---|
| | A | B | C |
| 1 | +10 | 35 | 23 |
| 2 | -10 | 35 | 23 |
| 3 | +38 | 35 | 23 |
| 4 | -38 | 35 | 23 |
| 5 | +40 | 36 | 25 |
| 6 | -40 | 36 | 25 |

Here the notations A, B, C stands for:-
A: - Angles of Alignment with respect to Y axis from FRAV3D database
B: - No of 3D images aligned against Y-axes
C: - No of alignments detected correctly by our algorithm
The fig 12 shows some sample taken from the FRAV3D database for images rotated about Z-axis.

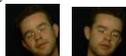

Fig 12. Some samples rotated about z axis from FRAV3D Database

The results of pose –alignment across Z axes are enlisted in Table III.

TABLE III

| Table 2 | Detection of Pose Alignment across Z axes | | |
|---|---|---|---|
| | A | B | C |
| 1 | +18 | 35 | 22 |
| 2 | -18 | 35 | 22 |
| 3 | +38 | 35 | 23 |
| 4 | -38 | 35 | 24 |
| 5 | +40 | 36 | 24 |
| 6 | -40 | 36 | 22 |

Here the notations A, B, C stands for:-
A: - Angles of Alignment with respect to Z axis from FRAV3D database
B: - No of 3D images aligned against Z-axes
C: - No of alignments detected correctly by our algorithm

## V. CONCLUSION

In conclusion, it must be said that the performance of our algorithm is an average one due to the present of outliers while calculating curvature values. Also for large pose variations, max intensity technique would not work satisfactorily. In future, we are trying to detect a method of how to discard outliers to a great extent. Since we have been successful in detecting the pose alignment of 3D faces, our next phase of work would be to detect orientations with respect to positive and negative X, Y and Z axes and also determine an approach of registration by rotation, scaling and translation parameter finding.